\documentclass[11pt,a4paper]{article}
\usepackage[hyperref]{acl2019}
\usepackage{latexsym}
\usepackage{todonotes}
\usepackage{algorithm,algpseudocode}
\usepackage{url}
\usepackage{booktabs}
\usepackage{siunitx}
\renewcommand{\th}[1]{\multicolumn{1}{c}{#1}}
\newcolumntype{B}{S[table-format=2.1,table-auto-round]} 
\newcolumntype{N}{S[table-format=2.1,table-space-text-post=M]} 
\newcolumntype{P}{S[table-format=3,table-space-text-post=\%]} 
\usepackage{multirow}
\usepackage{amsmath,mathtools}
\DeclareMathOperator*{\argmin}{arg\,min}
\usepackage{txfonts}
\usepackage{tikz}
\usepackage{paralist}

\aclfinalcopy 

\title{Auto-Sizing the Transformer Network: \\[2pt]  Improving Speed, Efficiency, and Performance \\[2pt] for Low-Resource Machine Translation}

\author{Kenton Murray \enskip Jeffery Kinnison \enskip Toan Q. Nguyen \enskip  Walter Scheirer \enskip David Chiang \\
Department of Computer Science and Engineering \\
  University of Notre Dame \\
  \text{\{kmurray4, jkinniso, tnguye28, walter.scheirer, dchiang\}@nd.edu} }

\date{}

\begin{document}
\maketitle
\begin{abstract}
  Neural sequence-to-sequence models, particularly the Transformer,
  are the state of the art in machine translation.
  Yet these neural networks are very
  sensitive to architecture and hyperparameter settings. 
  Optimizing these settings by grid or random search is computationally expensive because it requires many training runs. In this paper, we incorporate architecture search into a single training run through \emph{auto-sizing}, which uses regularization to delete neurons in a network over the course of training. On very low-resource language pairs, we show that auto-sizing can improve BLEU scores by up to 3.9 points while removing one-third of the parameters from the model. 
\end{abstract}

\section{Introduction}

Encoder-decoder based neural network models are the state-of-the-art in machine translation. However, these models are very dependent on selecting optimal hyperparameters and architectures. This problem is exacerbated in very low-resource data settings where the potential to overfit is high. 
Unfortunately, these searches are computationally expensive. For instance, \citet{britz2017massive} used over 250,000 GPU hours to compare various recurrent neural network based encoders and decoders for machine translation. \citet{strubell2019energy} demonstrated the neural architecture search for a large NLP model emits over four times the carbon dioxide relative to a car over its entire lifetime. 

Unfortunately, optimal settings are highly dependent on both the model and the task, which means that this process must be repeated often.
As a case in point, the Transformer architecture has become the best performing encoder-decoder model for machine translation \cite{vaswani2017attention}, displacing RNN-based models \cite{bahdanau2014neural} along with much conventional wisdom about how to train such models. Vaswani et al.~ran experiments varying numerous hyperparameters of the Transformer, but only on high-resource datasets among linguistically similar languages. \citet{popel2018training} explored ways to train Transformer networks, but only on a high-resource dataset in one language pair. Less work has been devoted to finding best practices for smaller datasets and linguistically divergent language pairs.

In this paper, we apply \emph{auto-sizing} \cite{murrayauto}, which is a type of architecture search conducted during training, to the Transformer. We show that it is effective on very low-resource datasets and can reduce model size significantly, while being substantially faster than other architecture search methods. 
We make three main contributions.
\begin{trivlist}
\item 1. We demonstrate the effectiveness of auto-sizing on the Transformer network by significantly reducing model size, even though the number of parameters in the Transformer is orders of magnitude larger than previous natural language processing applications of auto-sizing.
\item 2. We demonstrate the effectiveness of auto-sizing on translation quality in very low-resource settings. On four out of five language pairs, we obtain improvements in BLEU over a recommended low-resource baseline architecture. Furthermore, we are able to do so an order of magnitude faster than random search.
\item 3. We release GPU-enabled implementations of proximal operators used for auto-sizing. Previous authors \citep{boyd2011distributed,duchi2008efficient} have given efficient algorithms, but they don't necessarily parallelize well on GPUs. Our variations are optimized for GPUs and are implemented as a general toolkit and are released as open-source software.\footnote{https://github.com/KentonMurray/ProxGradPytorch}
\end{trivlist}

\section{Hyperparameter Search}

While the parameters of a neural network are optimized by gradient-based training methods, hyperparameters are values that are typically fixed before training begins, such as layer sizes and learning rates, and can strongly influence the outcome of training. 
Hyperparameter optimization is a search over the possible choices of hyperparameters for a neural network, with the objective of minimizing some cost function (e.g., error, time to convergence, etc.). Hyperparameters may be selected using a variety of methods, most often manual tuning, grid search \cite{duan2005best}, or random search \cite{bergstra2012random}. Other methods, such as Bayesian optimization \cite{bergstra2011algorithms,snoek2012practical}, genetic algorithms \cite{benardos2007optimizing,friedrichs2005evolutionary,vose2019}, and hypergradient updates \cite{maclaurin2015gradient}, attempt to direct the selection process based on the objective function. All of these methods require training a large number of networks with different hyperparameter settings.

In this work, we focus on a type of hyperparameter optimization called auto-sizing introduced by \citet{murrayauto} which only requires training one network once. Auto-sizing focuses on driving groups of weights in a parameter tensor to zero through regularization. \citet{murrayauto} focused on the narrow case of two hidden layers in a feed-forward neural network with a rectified linear unit activation.  In this work, we look at the broader case of all of the non-embedding parameter matrices in the encoder and decoder of the Transformer network. 

\section{GPU Optimized Proximal Gradient Descent}\label{sec:pgd}

\begin{algorithm}
\newcommand{\sorted}{\textit{sorted}}
\newcommand{\diff}{\delta}
\begin{algorithmic}[1]
\Require{Vector $\mathbf{v}$ with $n$ elements}
\Ensure{Decrease the largest absolute value in $\mathbf{v}$ until the total decrease is $\eta\lambda$}
\State{$v_i \leftarrow |v_i|$}
\State{sort $\mathbf{v}$ in decreasing order} \label{line:sort}
\State{$\diff_i \leftarrow v_i - v_{i+1}$, $\diff_n \leftarrow v_n$}
\State{$\displaystyle c_i \leftarrow \sum_{i'=1}^i i'\diff_{i'}$} \Comment{prefix sum} \label{line:sum1}
\State{$b_i = \frac1i(\text{clip}_{[c_{i-1},c_i]} (\eta\lambda) - c_{i-1})$}
\State{$\displaystyle p_i = \sum_{i'=i}^n b_{i'}$} \Comment{suffix sum} \label{line:sum2}
\State{$\mathbf{v} \leftarrow \mathbf{v} - \mathbf{p}$}
\State{restore order and signs of $\mathbf{v}$}
\caption{Parallel $\ell_{\infty}$ proximal step}
\label{gpu}
\end{algorithmic}
\end{algorithm}

\citet{murrayauto} train a neural network while using a regularizer to prune units from the network, minimizing:
\begin{equation*}
\mathcal{L} = -\sum_{\text{$f, e$ in data}} \log P(e \mid f; W) + \lambda R(\|W\|),
\end{equation*}
where $W$ are the parameters of the model and $R$ is a regularizer. For simplicity, assume that the parameters form a single matrix $W$ of weights. \citet{murrayauto} try two regularizers:
\begin{align*}
R(W) &= \sum_i \left(\sum_j W_{ij}^2\right)^{\frac12} && (\ell_{2,1}) \\
R(W) &= \sum_i \max_j |W_{ij}| && (\ell_{\infty,1})
\end{align*}
The optimization is done using proximal gradient descent \citep{parikh+boyd:2014}, which alternates between stochastic gradient descent steps and proximal steps:
\begin{align*}
W &\leftarrow W - \eta \nabla \log P(e \mid f; w) \\
W &\leftarrow \argmin_{W'} \left(\frac1{2\eta} \|W-W'\|^2 + R(W') \right)
\end{align*}
To perform the proximal step for the $\ell_{\infty,1}$ norm, they rely on a quickselect-like algorithm that runs in $O(n)$ time \citep{duchi2008efficient}. However, this algorithm does not parallelize well. Instead, we use Algorithm \ref{gpu}, which is similar to that of \citet{quattoni2009efficient}, on each row of $W$.

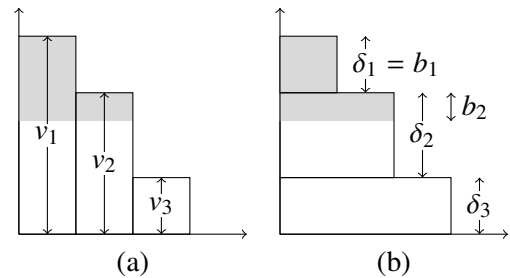
\begin{figure}
\tikzset{inner sep=1pt}
\begin{center}
\begin{tabular}{cc}
\begin{tikzpicture}[scale=0.75]
\draw[->] (0,0)--(0,4);
\draw[->] (0,0)--(4,0);

\draw[fill=gray!30,draw=none] (0,2) rectangle (2, 2.5);
\draw[fill=gray!30,draw=none] (0,2.5) rectangle (1,3.5);

\draw (0,0) rectangle (1,3.5);
\draw[<->] (0.5,0) to node[fill=white] {$v_1$} (0.5,3.5);


\draw (1,0) rectangle (2,2.5);
\draw[<->] (1.5,0) to node[fill=white] {$v_2$} (1.5,2.5);


\draw (2,0) rectangle (3,1);
\draw[<->] (2.5,0) to node[fill=white] {$v_3$} (2.5,1);


\end{tikzpicture}
&
\begin{tikzpicture}[scale=0.75]
\draw[->] (0,0)--(0,4);
\draw[->] (0,0)--(4,0);

\draw[fill=gray!30,draw=none] (0,2) rectangle (2, 2.5);
\draw[fill=gray!30] (0,2.5) rectangle (1,3.5);

\draw (0,0) rectangle (3,1);
\draw (0,1) rectangle (2,2.5);

\draw[<->] (1.5,2.5) to node[fill=white] {$\delta_1\rlap{${}= b_1$}$} (1.5,3.5);

\draw[<->] (2.5,1) to node[fill=white] {$\delta_2$} (2.5,2.5);

\draw[<->] (3,2) to node[auto=right,outer xsep=3pt] {$b_2$} (3,2.5);

\draw[<->] (3.5,0) to node[fill=white] {$\delta_3$} (3.5,1);

\end{tikzpicture}
 \\
 (a) & (b)

\end{tabular}
\end{center}
\caption{Illustration of Algorithm~\ref{gpu}. The shaded area, here with value $\eta\lambda = 2$, represents how much the $\ell_{\infty}$ proximal step will remove from a sorted vector.} \label{fig:proximal}
\end{figure}

The algorithm starts by taking the absolute value of each entry and sorting the entries in decreasing order. Figure~\ref{fig:proximal}a shows a histogram of sorted absolute values of an example $\mathbf{v}$. Intuitively, the goal of the algorithm is to cut a piece off the top with area $\eta\lambda$ (in the figure, shaded gray). 

We can also imagine the same shape as a stack of horizontal layers (Figure~\ref{fig:proximal}b), each $i$ wide and $\delta_i$ high, with area $i\delta_i$; then $c_i$ is the cumulative area of the top $i$ layers. This view makes it easier to compute where the cutoff should be. 
Let $k$ be the index such that $\eta\lambda$ lies between $c_{k-1}$ and $c_k$. Then $b_i = \delta_i$ for $i < k$; $b_k = \frac1k(\eta\lambda - c_{k-1})$; and $b_i = 0$ for $i > k$. In other words, $b_i$ is how much height of the $i$th layer should be cut off. 

Finally, returning to Figure~\ref{fig:proximal}b, $p_i$ is the amount by which $v_i$ should be decreased (the height of the gray bars). (The vector $\mathbf{p}$ also happens to be the projection of $\mathbf{v}$ onto the $\ell_1$ ball of radius $\eta\lambda$.)

Although this algorithm is less efficient than the quickselect-like algorithm when run in serial, the sort in line \ref{line:sort} and the cumulative sums in lines \ref{line:sum1} and \ref{line:sum2} \cite{ladner+fischer:jacm80} can be parallelized to run in $O(\log n)$ passes each.

\section{Transformer}

The Transformer network, introduced by \citet{vaswani2017attention}, is a sequence-to-sequence model in which both the encoder and the decoder consist of stacked self-attention layers. Each layer of the decoder can attend to the previous layer of the decoder and the output of the encoder.  The multi-head attention uses two affine transformations, followed by a softmax.
Additionally, each layer has a position-wise feed-forward neural network (FFN) with a hidden layer of rectified linear units:
\vspace{1mm}
\begin{equation*}
\text{FFN}(x) = W_2(\max(0,W_1x + b_1)) + b_2. 
\label{eq:ffn}
\end{equation*}
The hidden layer size (number of columns of $W_1$) is typically four times the size of the model dimension. Both the multi-head attention and the feed-forward neural network have residual connections that allow information to bypass those layers.

\subsection{Auto-sizing Transformer}
Though the Transformer has demonstrated remarkable success on a variety of datasets, it is highly over-parameterized. For example, the English-German WMT '14 Transformer-base model proposed in \citet{vaswani2017attention} has more than 60M parameters. Whereas early NMT models such as \citet{sutskever2014sequence} have most of their parameters in the embedding layers, the added complexity of the Transformer, plus parallel developments reducing vocabulary size \cite{sennrich-haddow-birch:2016:P16-12} and sharing embeddings \cite{press-wolf:2017:EACLshort} has shifted the balance. Nearly 31\% of the English-German Transformer's parameters are in the attention layers and 41\% in the position-wise feed-forward layers.

\begin{figure}
    \centering
    \includegraphics[width=5cm]{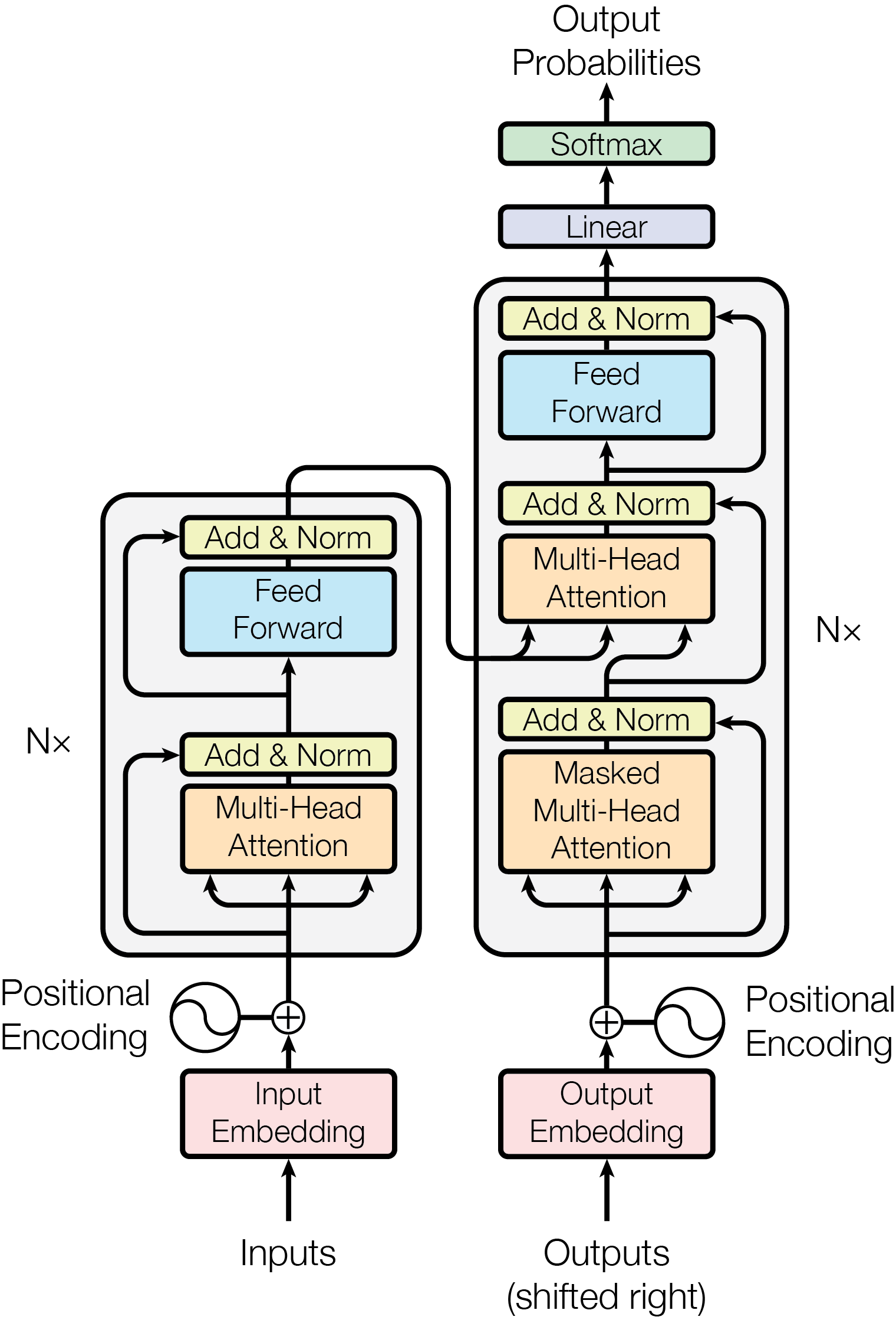}
    \caption{Architecture of the Transformer  \citep{vaswani2017attention}. We apply the auto-sizing method to the feed-forward (blue rectangles) and multi-head attention (orange rectangles) in all $n$ layers of the encoder and decoder. Note that there are residual connections that can allow information and gradients to bypass any layer we are auto-sizing.}
    \label{fig:transformer}
\end{figure}

Accordingly, we apply the auto-sizing method to the Transformer network, and in particular to the two largest components, the feed-forward layers and the multi-head attentions (blue and orange rectangles in Figure \ref{fig:transformer}). A difference from the work of \citet{murrayauto} is that there are residual connections that allow information to bypass the layers we are auto-sizing. If the regularizer drives all the neurons in a layer to zero, information can still pass through. Thus, auto-sizing can effectively prune out an entire layer.

\subsection{Random Search}

As an alternative to grid-based searches, random hyperparameter search has been demonstrated to be a strong baseline for neural network architecture searches as it can search between grid points to increase the size of the search space \cite{bergstra2012random}. In fact, \citet{li2019random} recently demonstrated that many architecture search methods do not beat a random baseline. In practice, randomly searching hyperparameter domains allows for an intuitive mixture of continuous and categorical hyperparameters with no constraints on differentiability~\cite{maclaurin2015gradient} or need to cast hyperparameter values into a single high-dimensional space to predict new values~\cite{bergstra2011algorithms}.

\section{Experiments}

\begin{table}
    \centering
    \begin{tabular}{l|S[table-format=3,table-space-text-post=M]}
         
         \toprule 
         Dataset & \th{Size} \\
         \midrule
         Ara--Eng & 234k \\
         Fra--Eng & 235k \\
         Hau--Eng & 45k \\
         Tir--Eng & 15k \\
         \bottomrule
    \end{tabular}
    \caption{Number of parallel sentences in training bitexts. The French-English and Arabic-English data is from the 2017 IWSLT campaign \cite{mauro2012wit3}. The much smaller Hausa-English and Tigrinya-English data is from the LORELEI project.}
    \label{tab:data}
\end{table}

All of our models are trained using the fairseq implementation of the Transformer \cite{gehring2017convs2s}.\footnote{https://github.com/pytorch/fairseq} Our GPU-optimized, proximal gradient algorithms are implemented in PyTorch and are publicly available.\footnote{https://github.com/KentonMurray/ProxGradPytorch} For the random hyperparameter search experiments, we use SHADHO,\footnote{https://github.com/jeffkinnison/shadho} which defines the hyperparameter tree, generates from it, and manages distributed resources \cite{kinnison2018shadho}. Our SHADHO driver file and modifications to fairseq are also publicly available.\footnote{https://bitbucket.org/KentonMurray/fairseq\_autosizing}

\subsection{Settings}

We looked at four different low-resource language pairs, running experiments in five directions: Arabic-English, English-Arabic, French-English, Hausa-English, and Tigrinya-English. The Arabic and French data comes from the IWSLT 2017 Evaluation Campaign \cite{mauro2012wit3}. The Hausa and Tigrinya data were provided by the LORELEI project with custom train/dev/test splits. For all languages, we tokenized and truecased the data using scripts from Moses \cite{koehn2007moses}. For the Arabic systems, we transliterated the data using the Buckwalter transliteration scheme. All of our systems were run using subword units (BPE) with 16,000 merge operations on concatenated source and target training data \cite{sennrich2016linguistic}. We clip norms at 0.1, use label smoothed cross-entropy with value 0.1, and an early stopping criterion when the learning rate is smaller than $10^{-5}$. All of our experiments were done using the Adam optimizer \cite{kingma2014adam}, a learning rate of $10^{-4}$, and dropout of 0.1. At test time, we decoded using a beam of 5 with length normalization \cite{boulanger2013audio} and evaluate using case-sensitive, detokenized BLEU \cite{papineni2002bleu}.

\subsubsection{Baseline} The originally proposed Transformer model is too large for our data size -- the model will overfit the training data. Instead, we use the recommended settings in fairseq for IWSLT German-English as a baseline since two out of our four language pairs are also from IWSLT. This architecture has 6 layers in both the encoder and decoder, each with 4 attention heads. Our model dimension is $d_{model}=512$, and our FFN dimension is 1024. 

\subsubsection{Auto-sizing parameters} Auto-sizing is implemented as two different types of group regularizers, $\ell_{2,1}$ and $\ell_{\infty,1}$. We apply the regularizers to the feed-forward network and multi-head attention in each layer of the encoder and decoder. We experiment across a range of regularization coefficient values, $\lambda$, that control how large the regularization proximal gradient step will be. We note that different regularization coefficient values are suited for different types or regularizers. Additionally, all of our experiments use the same batch size, which is also related to $\lambda$.

\subsubsection{Random search parameters}

As originally proposed, the Transformer network has 6 encoder layers, all identical, and 6 decoder layers, also identical. For our random search experiments, 
we sample the number of attention heads from $\{4, 8, 16\}$ and the model dimension ($d_{model}$) from $\{128, 256, 512, 1024, 2048\}$. Diverging from most implementations of the Transformer, we do not require the same number of encoder and decoder layers, but instead sample each from $\{2, 4, 6, 8\}$. Within a layer, we also sample the size of the feed-forward network (FFN), varying our samples over $\{512, 1024, 2048\}$. This too differs from most Transformer implementations, which have identical layer hyperparameters. 

\begin{table*}
    \centering
    \begin{tabular}{l|l|BNN}
    \toprule
        Language Pair & Search Strategy & \th{BLEU} & \th{Model Size} & \th{Training Time}   \\
        \midrule
        \multirow{4}{*}{Tir--Eng} & Standard Baseline & 3.64 & 39.6M & 1.2k \\
        & Random Search & 6.73 & 240.4M & 43.7k \\
        & Auto-sizing $\ell_{\infty,1}$ & 7.5 & 27.1M & 2.1k \\
        & Auto-sizing $\ell_{2,1}$ & 7.4 & 27.1M & 1.2k \\
        \midrule
        \multirow{4}{*}{Hau--Eng} & Standard Baseline & 13.85 & 39.6M & 4.2k \\
        & Random Search & 17.16 & 15.4M & 87.0k \\
        & Auto-sizing $\ell_{\infty,1}$ & 15.00 & 27.1M & 7.6k \\
        & Auto-sizing $\ell_{2,1}$ & 14.82 & 27.1M & 3.5k \\
        \midrule
        \multirow{4}{*}{Fra--Eng} & Standard Baseline & 35.0 & 39.6M & 11.3k \\
        & Random Search & 35.3 & 121.2M & 116.0k \\
        & Auto-sizing $\ell_{\infty,1}$ & 34.3 & 27.1M & 28.8k \\
        & Auto-sizing $\ell_{2,1}$ & 33.5 & 27.1M & 11.5k \\
        \bottomrule
        
    \end{tabular}
    \caption{Comparison of BLEU scores, model size, and training time on Tigrinya-English, Hausa-English, and French-English. Model size is the total number of parameters. Training time is measured in seconds. Baseline is the recommended low-resource architecture in fairseq. Random search represents the best model found from 72 (Tigrinya), 40 (Hausa), and 10 (French) different randomly generated architecture hyperparameters. Both auto-sizing methods, on both languages, start with the exact same initialization and number of parameters as the baseline, but converge to much smaller models across all language pairs. On the very low-resource languages of Hausa and Tigrinya auto-sizing finds models with better BLEU scores. Random search is eventually able to find better models on French and Hausa, but is an order of magnitude slower.}
    \label{tab:random_vs_ti_hau}
\end{table*}

\subsection{Auto-sizing vs. Random Search}

Table \ref{tab:random_vs_ti_hau} compares the performance of random search with auto-sizing across, BLEU scores, model size, and training times. The baseline system, the recommended IWSLT setting in fairseq, has almost 40 million parameters. Auto-sizing the feed-forward network sub-components in each layer of this baseline model with $\ell_{2,1}=10.0$ or $\ell_{\infty,1}=100.0$ removes almost one-third of the total parameters from the model. For Hausa-English and Tigrinya-English, this also results in substantial BLEU score gains, while only slightly hurting performance for French-English. The BLEU scores for random search beats the baseline for all language pairs, but auto-sizing still performs best on Tigrinya-English -- even with 72 different, random hyperparameter configurations.

Auto-sizing trains in a similar amount of time to the baseline system, whereas the cumulative training time for all of the models in random search is substantially slower. Furthermore, for Tigrinya-English and French-English, random search found models that were almost 10 and 5 times larger respectively than the auto-sized models.

\subsection{Training times}

One of the biggest downsides of searching over architectures using a random search process is that it is very time and resource expensive. Contrary to that, auto-sizing relies on only training \textit{one} model.

Auto-sizing relies on a proximal gradient step after a standard gradient descent step. However, the addition of these steps for our two group regularizers does not significantly impact training times. Table \ref{tab:times} shows the total training time for both $\ell_{2,1}=0.1$ and $\ell_{\infty,1}=0.5$. Even with the extra proximal step, auto-sizing using $\ell_{2,1}$ actually converges faster on two of the five language pairs. Note that these times are for smaller regularization coefficients. Larger coefficients will cause more values to go to zero, which will make the model converge faster.

\begin{table}
\centering
\begin{tabular}{l|NNN}
\toprule
    Language Pair & \th{Baseline} & \th{$\ell_{2,1}$} & \th{$\ell_{\infty,1}$}  \\
    \midrule
    Fra--Eng & 11.3k & 11.5k & 28.8k \\ 
    Ara--Eng & 15.1k & 16.6k & 40.8k \\ 
    Eng--Ara & 16.6k & 11.0k & 21.9k \\ 
    Hau--Eng & 4.2k & 3.5k & 7.6k \\ 
    Tir--Eng & 1.2k & 1.2k & 2.1k \\ 
    \bottomrule
\end{tabular}
\caption{Overall training times in seconds on a Nvidia GeForce GTX 1080Ti GPU for small regularization values. Note that high regularization values will delete too many values and cause training to end sooner. In general, $\ell_{2,1}$ regularization does not appreciably slow down training, but $\ell_{\infty,1}$ can be twice as slow. Per epoch, roughly the same ratios in training times hold.}
\label{tab:times}
\end{table}

\subsection{Auto-sizing Sub-Components}

As seen above, on very low-resource data, auto-sizing is able to quickly learn smaller, yet better, models than the recommended low-resource transformer architecture.
Here, we look at the impact of applying auto-sizing to various sub-components of the Transformer network. In section \ref{sec:pgd}, following the work of \citet{murrayauto}, auto-sizing is described as
intelligently applying a group regularizer to our objective function. The relative weight, or regularization coefficient, is a hyperparameter defined as $\lambda$. In this section, we also look at the impact of varying the strength of this regularization coefficient.

Tables \ref{tab:l21_fc_all} and \ref{tab:linf1_fc_all} demonstrate the impact of varying the regularization coefficient strength has on BLEU scores and model size across various model sub-components. Recall that each layer of the Transformer network has multi-head attention sub-components and a feed-forward network sub-component. We denote experiments only applying auto-sizing to feed-forward network as ``FFN''. We also experiment with auto-sizing the multi-head attention in conjunction with the FFN, which we denote ``All''. A regularization coefficient of 0.0 refers to the baseline model without any auto-sizing. Columns which contain percentages refer to the number of rows in a PyTorch parameter that auto-sizing was applied to, that were entirely driven to zero. In effect, neurons deleted from the model. Note that individual values in a row may be zero, but if even a single value remains, information can continue to flow through this and it is not counted as deleted. Furthermore, percentages refer only to the parameters that auto-sizing was applied to, not the entire model. As such, with the prevalence of residual connections, a value of 100\% does not mean the entire model was deleted, but merely specific parameter matrices. More specific experimental conditions are described below.

\begin{table*}
    \centering
    \scalebox{0.99}{%
    \begin{tabular}{@{}l|l|B|B|BP|BP|BP|BP@{}}
    \toprule
    & & \multicolumn{6}{c}{$\ell_{2,1}$ coefficient} \\
     & Model Portion & 0.0 & 0.1 & \multicolumn{2}{c|}{0.25} & \multicolumn{2}{c|}{0.5} & \multicolumn{2}{c|}{1.0} & \multicolumn{2}{c}{10.0}  \\
    \midrule
    \multirow{6}{*}{Hau--Eng} & Encoder All & 
    13.85
    & 15.98 & 17.06 && 17.44 && 15.32 & 89\% & 16.43 & 100\%  \\
    & Encoder FFN & & 15.42 & 15.06 && 16.26 && 15.91 & 100\% & 16.66 & 100\%  \\
    & Decoder All & & 12.55 & 16.05 && 16.18 && 13.03 & 3\% & 0.00 & 63\% \\
    & Decoder FFN & & 11.84 & 14.65 && 14.39 && 11.65 & 79\% & 13.11 & 100\% \\
    & Enc+Dec All & & 15.77 & 17.39 && 17.80 && 12.50 & 61\% & 0.00 & 100\% \\
    & Enc+Dec FFN & & 14.69 & 15.27 && 14.24 && 12.75 & 86\% & 14.82 & 100\% \\
    \midrule
    \multirow{3}{*}{Tir--Eng} & Encoder All & 
    3.64
    & 3.25 & 4.68 && 5.28 && 7.23 & & 8.43 & 100\%  \\
    & Enc+Dec All & & 3.83 & 4.00 && 6.47 && 6.97 & & 0.00 & 100\% \\
    & Enc+Dec FFN & & 4.01 & 4.20 && 3.32 && 5.07 & & 7.39 & 100\% \\
    \midrule
    \multirow{3}{*}{Fra--Eng} & Encoder All & 
    34.95
    & 35.70 & 34.47 && 34.14 && 33.57 & 97\% & 32.79 & 100\% \\
    & Enc+Dec All & & 35.15 & 33.10 && 29.79 & 23\% & 24.15 & 73\% & 0.28 & 100\% \\
    & Enc+Dec FFN & & 35.59 & 34.99 && 34.15 & 15\% & 34.20 & 98\% & 33.49 & 100\% \\
    \midrule
    \multirow{2}{*}{Ara--Eng} & Enc+Dec All & 
    27.91
    & 28.01 & 24.74 & 1\% & 20.90 & 20\% & 14.31 & 72\% & 0.28 & 100\% \\
    & Enc+Dec FFN & & 26.89 & 26.72 & 1\% & 25.50 & 23\% & 25.86 & 97\% & 25.73 & 100\% \\
    \midrule
    \multirow{2}{*}{Eng--Ara} & Enc+Dec All & 
    9.41
    & 8.70 & 7.48 && 5.81 &  23\% & 3.73 & 73\% & 0.00 & 100\% \\
    & Enc+Dec FFN & & 8.63 & 8.31 & 3\% & 8.27 & 22\% & 7.91 & 93\% & 8.03 & 100\%  \\
    \bottomrule
    \end{tabular}}
    \caption{BLEU scores and percentage of parameter rows deleted by auto-sizing on various sub-components of the model, across varying strengths of $\ell_{2,1}$ regularization. 0.0 refers to the baseline without any regularizer. Blank spaces mean less than 1\% of parameters were deleted. In the two very low-resource language pairs (Hausa-English and Tigrinya-English), deleting large portions of the encoder can actually help performance. However, deleting the decoder hurts performance.}
    \label{tab:l21_fc_all}
    
\end{table*}


\begin{table*}
    \centering
    \begin{tabular}{l|l|B|B|B|B|B|BP|BP}
    \toprule
    && \multicolumn{9}{c}{$\ell_{\infty,1}$} \\
    & & 0.0 & 0.1 & {0.25} & 0.5 & 1.0 & \multicolumn{2}{c|}{10.0} &
    \multicolumn{2}{c}{100.0}  \\
    \midrule
    \multirow{2}{*}{Hau--Eng} & Enc+Dec All & 
    13.85 & 15.50 & 14.73 & 15.99 & 16.66 & 14.89 & 4\% & 1.45 & 100\%\\
    & Enc+Dec FFN & & 13.43 & 14.34 & 14.13 & 12.89 & 15.27 & 0\% & 15.00 & 100\% \\
    \midrule
    \multirow{2}{*}{Tir--Eng} & Enc+Dec All & 
    3.64 & 4.55 & 3.35 & 3.39 & 3.71 & 7.35 & 0\% & 2.41 & 100\% \\
    & Enc+Dec FFN & & 3.56 & 3.82 & 3.92 & 3.59 & 4.68 & 0\% & 7.53 & 100\% \\
    \midrule
    \multirow{2}{*}{Fra--Eng} & Enc+Dec All & 
    34.95 & 35.20 & 35.39 & 34.93 & 35.29 & 26.26 & 13\%  & 1.73 & 100\%\\
    & Enc+Dec FFN & & 34.77 & 35.50 & 35.40 & 35.03 & 34.12 & 0\% & 34.26 & 100\% \\
    \midrule
    \multirow{2}{*}{Ara--Eng} & Enc+Dec All & 
    27.91
    & 27.31 & 27.50 & 27.57 & 26.92 & 18.51 & 22\% & 0.56 & 100\% \\
    & Enc+Dec FFN & & 27.83 & 27.17 & 28.31 & 27.63 & 25.42 & 0\% & 25.35 & 100\% \\
    \midrule
    \multirow{2}{*}{Eng--Ara} & Enc+Dec All &
    9.41
    & 9.13 & 8.34 & 8.38 & 8.66 & 5.16 & 25\%  & 0.56 & 100\% \\
    & Enc+Dec FFN & & 8.81 & 9.17 & 9.03 & 8.88 & 8.23 & 0\% & 8.25 & 100\% \\
 \bottomrule
    
    \end{tabular}
    \caption{BLEU scores and percentage of model deleted using auto-sizing with various $l_{\infty,1}$ regularization strengths. On the very low-resource language pairs of Hausa-English and Tigrinya-English, auto-sizing the feed-forward networks of the encoder and decoder can improve BLEU scores.}
    \label{tab:linf1_fc_all}
\end{table*}

\subsubsection{FFN matrices and multi-head attention}

Rows corresponding to ``All'' in tables \ref{tab:l21_fc_all} and \ref{tab:linf1_fc_all} look at the impact of varying the strength of both the $\ell_{\infty,1}$ and $\ell_{2,1}$ regularizers across all learned parameters in the encoder and decoders (multi-head and feed-forward network parameters). Using $\ell_{\infty,1}$ regularization (table \ref{tab:linf1_fc_all}), auto-sizing beats the baseline BLEU scores on three language pairs: Hau--Eng, Tir--Eng, Fra--Eng. However, BLEU score improvements only occur on smaller regularization coefficients that do not delete model portions.

Looking at $\ell_{2,1}$ regularization across all learned parameters of both the encoder and decoder (``Enc+Dec All'' in table \ref{tab:l21_fc_all}), auto-sizing beats the baseline on four of the five language pairs (all except Eng--Ara).
Again, BLEU gains are on smaller regularization coefficients, and stronger regularizers that delete parts of the model hurt translation quality.
Multi-head attention is an integral portion of the Transformer model and auto-sizing this generally leads to performance hits.

\subsubsection{FFN matrices}

As the multi-head attention is a key part of the Transformer, we also looked at auto-sizing just the feed-forward sub-component in each layer of the encoder and decoder.
Rows deonted by ``FFN'' in tables \ref{tab:l21_fc_all} and \ref{tab:linf1_fc_all} look at applying auto-sizing to all of the feed-forward network sub-components of the Transformer, but not to the multi-head attention. With $\ell_{\infty,1}$ regularization, we see BLEU improvements on four of the five language pairs. For both Hausa-English and Tigrinya-English, we see improvements even after deleting all of the feed-forward networks in all layers. Again, the residual connections allow information to flow around these sub-components. Using $\ell_{2,1}$ regularization, we see BLEU improvements on three of the language pairs. Hausa-English and Tigrinya-English maintain a BLEU gain even when deleting all of the feed-forward networks.

Auto-sizing only the feed-forward sub-component, and not the multi-head attention part, results in better BLEU scores,
even when deleting all of the feed-forward network components.
Impressively, this is with a model that has fully \textit{one-third} fewer parameters in the encoder and decoder layers. This is beneficial for faster inference times and smaller disk space.

\subsubsection{Encoder vs. Decoder}

In table \ref{tab:l21_fc_all}, experiments on Hau-Eng look at the impact of auto-sizing either the encoder or the decoder separately. Applying a strong enough regularizer to delete portions of the model ($\ell_{2,1} \ge 1.0$) only to the decoder (``Decoder All'' and ``Decoder FFN'') results in a BLEU score drop. However, applying auto-sizing to only the encoder (``Encoder All'' and ``Encoder FFN'') yields a BLEU gain while creating a smaller model. Intuitively, this makes sense as the decoder is closer to the output of the network and requires more modeling expressivity.

In addition to Hau--Eng, table \ref{tab:l21_fc_all} also contains experiments looking at auto-sizing all sub-components of all encoder layers of Tir--Eng and Fra--Eng. For all three language pairs, a small regularization coefficient for the $\ell_{2,1}$ regularizer applied to the encoder increases BLEU scores. However, no rows are driven to zero and the model size remains the same. Consistent with Hau--Eng, using a larger regularization coefficient drives all of the encoder's weights to all zeros. For the smaller Hau--Eng and Tir--Eng datasets, this actually results in BLEU gains over the baseline system. Surprisingly, even on the Fra--Eng dataset, which has more than 15x as much data as Tir--Eng, the performance hit of deleting the \textit{entire} encoder was only 2 BLEU points.

Recall from Figure \ref{fig:transformer} that there are residual connections that allow information and gradients to flow around both the multi-head attention and feed-forward portions of the model. Here, we have the case that all layers of the encoder have been completely deleted. However, the decoder still attends over the source word and positional embeddings due to the residual connections. We hypothesize that for these smaller datasets that there are too many parameters in the baseline model and over-fitting is an issue.

\subsection{Random Search plus Auto-sizing}

Above, we have demonstrated that auto-sizing is able to learn smaller models, faster than random search, often with higher BLEU scores. To compare whether the two architecture search algorithms (random and auto-sizing) can be used in conjunction, we also looked at applying both $\ell_{2,1}$ and $\ell_{\infty,1}$ regularization techniques to the FFN networks in all encoder and decoder layers \textit{during} random search. In addition, this looks at how robust the auto-sizing method is to different initial conditions.




 For a given set of hyperparameters generated by the random search process, we initialize three \textit{identical} models and train a baseline as well as one with each regularizer ($\ell_{2,1}=1.0$ and $\ell_{\infty,1}=10.0$). We trained 216 Tir--Eng models ($3\cdot72$ hyperparameter config.), 120 Hau--Eng, 45 Ara--Eng, 45 Eng--Ara, and 30 Fra--Eng models. Using the model with the best dev perplexity found during training, table \ref{tab:shadho_autosizing} shows the test BLEU scores for each of the five language pairs. For the very low-resource language pairs of Hau--Eng and Tir--Eng, auto-sizing is able to find the best BLEU scores.

\begin{table}
    \centering
    \begin{tabular}{l|r|r|r}
    \toprule
   & none & $\ell_{2,1}$ & $\ell_{\infty,1}$  \\
    \midrule
    Hau--Eng & 17.2 & 16.6 & 17.8 \\
    Tir--Eng & 6.7 & 7.9 & 7.6 \\
    Fra--Eng & 35.4 & 34.7 & 34.1  \\
    Ara--Eng & 27.6 & 25.6 & 25.9 \\
    Eng--Ara & 9.0 & 7.6 & 8.4 \\
\bottomrule
    \end{tabular}
    \caption{Test BLEU scores for the models with the best dev perplexity found using random search over number of layers and size of layers. Regularization values of $\ell_{2,1}=1.0$ and $\ell_{\infty,1}=10.0$ were chosen based on tables \ref{tab:l21_fc_all} and \ref{tab:linf1_fc_all} as they encouraged neurons to be deleted. For the very low-resource language pairs, auto-sizing helped in conjunction with random search.}
    \label{tab:shadho_autosizing}
    \vspace{-4mm}
\end{table}

\section{Conclusion}

In this paper, we have demonstrated the effectiveness of auto-sizing on the Transformer network. On very low-resource datasets, auto-sizing was able to improve BLEU scores by up to 3.9 points while simultaneously deleting one-third of the parameters in the encoder and decoder layers. This was accomplished while being significantly faster than other search methods.

Additionally, we demonstrated how to apply proximal gradient methods efficiently using a GPU. Previous work on optimizing proximal gradient algorithms serious impacts speed performance when the computations are moved off of a CPU and parallelized. Leveraging sorting and prefix summation, we reformulated these methods to be GPU efficient.

Overall, this paper has demonstrated the efficacy of auto-sizing on a natural language processing application with orders of magnitude more parameters than previous work. With a focus on speedy architecture search and an emphasis on optimized GPU algorithms, auto-sizing is able to improve machine translation on very low-resource language pairs without being resource or time-consuming.

\section*{Acknowledgements}
This research was supported in part by University of Southern California, subcontract 67108176 under DARPA contract HR0011-15-C-0115. We would like to thank Justin DeBenedetto for helpful comments.

\bibliography{autosizing}

\begin{thebibliography}{31}
\expandafter\ifx\csname natexlab\endcsname\relax\def\natexlab#1{#1}\fi

\bibitem[{Bahdanau et~al.(2015)Bahdanau, Cho, and Bengio}]{bahdanau2014neural}
Dzmitry Bahdanau, Kyunghyun Cho, and Yoshua Bengio. 2015.
\newblock Neural machine translation by jointly learning to align and
  translate.
\newblock In \emph{Proc. ICLR}.

\bibitem[{Benardos and Vosniakos(2007)}]{benardos2007optimizing}
P.~G. Benardos and G.-C. Vosniakos. 2007.
\newblock Optimizing feedforward artificial neural network architecture.
\newblock \emph{Engineering Applications of Artificial Intelligence},
  20:365--382.

\bibitem[{Bergstra and Bengio(2012)}]{bergstra2012random}
James Bergstra and Yoshua Bengio. 2012.
\newblock Random search for hyper-parameter optimization.
\newblock \emph{Journal of Machine Learning Research}, 13:281--305.

\bibitem[{Bergstra et~al.(2011)Bergstra, Bardenet, Bengio, and
  K{\'e}gl}]{bergstra2011algorithms}
James~S. Bergstra, R{\'e}mi Bardenet, Yoshua Bengio, and Bal{\'a}zs K{\'e}gl.
  2011.
\newblock Algorithms for hyper-parameter optimization.
\newblock In \emph{Advances in Neural Information Processing Systems}, pages
  2546--2554.

\bibitem[{Boulanger-Lewandowski et~al.(2013)Boulanger-Lewandowski, Bengio, and
  Vincent}]{boulanger2013audio}
Nicolas Boulanger-Lewandowski, Yoshua Bengio, and Pascal Vincent. 2013.
\newblock Audio chord recognition with recurrent neural networks.
\newblock In \emph{Proc. International Society for Music Information
  Retrieval}, pages 335--340.

\bibitem[{Boyd et~al.(2010)Boyd, Parikh, Chu, Peleato, and
  Eckstein}]{boyd2011distributed}
Stephen Boyd, Neal Parikh, Eric Chu, Borja Peleato, and Jonathan Eckstein.
  2010.
\newblock Distributed optimization and statistical learning via the alternating
  direction method of multipliers.
\newblock \emph{Foundations and Trends in Machine learning}, 3(1):1--122.

\bibitem[{Britz et~al.(2017)Britz, Goldie, Luong, and Le}]{britz2017massive}
Denny Britz, Anna Goldie, Minh-Thang Luong, and Quoc Le. 2017.
\newblock Massive exploration of neural machine translation architectures.
\newblock In \emph{Proc. EMNLP}, pages 1442--1451.

\bibitem[{Duan and Keerthi(2005)}]{duan2005best}
Kai-Bo Duan and S.~Sathiya Keerthi. 2005.
\newblock Which is the best multiclass {SVM} method? {A}n empirical study.
\newblock In \emph{International Workshop on Multiple Classifier Systems},
  pages 278--285.

\bibitem[{Duchi et~al.(2008)Duchi, Shalev-Shwartz, Singer, and
  Chandra}]{duchi2008efficient}
John Duchi, Shai Shalev-Shwartz, Yoram Singer, and Tushar Chandra. 2008.
\newblock Efficient projections onto the {$\ell_1$}-ball for learning in high
  dimensions.
\newblock In \emph{Proc. ICML}, pages 272--279.

\bibitem[{Friedrichs and Igel(2005)}]{friedrichs2005evolutionary}
Frauke Friedrichs and Christian Igel. 2005.
\newblock Evolutionary tuning of multiple {SVM} parameters.
\newblock \emph{Neurocomputing}, 64:107--117.

\bibitem[{Gehring et~al.(2017)Gehring, Auli, Grangier, Yarats, and
  Dauphin}]{gehring2017convs2s}
Jonas Gehring, Michael Auli, David Grangier, Denis Yarats, and Yann~N. Dauphin.
  2017.
\newblock {Convolutional Sequence to Sequence Learning}.
\newblock In \emph{Proc. ICML}.

\bibitem[{Kingma and Ba(2015)}]{kingma2014adam}
Diederik~P. Kingma and Jimmy~Lei Ba. 2015.
\newblock {Adam}: A method for stochastic optimization.
\newblock In \emph{Proc. ICLR}.

\bibitem[{Kinnison et~al.(2018)Kinnison, Kremer-Herman, Thain, and
  Scheirer}]{kinnison2018shadho}
Jeffery Kinnison, Nathaniel Kremer-Herman, Douglas Thain, and Walter Scheirer.
  2018.
\newblock Shadho: Massively scalable hardware-aware distributed hyperparameter
  optimization.
\newblock In \emph{Proc. IEEE Winter Conference on Applications of Computer
  Vision (WACV)}, pages 738--747.

\bibitem[{Koehn et~al.(2007)Koehn, Hoang, Birch, Callison-Burch, Federico,
  Bertoldi, Cowan, Shen, Moran, Zens et~al.}]{koehn2007moses}
Philipp Koehn, Hieu Hoang, Alexandra Birch, Chris Callison-Burch, Marcello
  Federico, Nicola Bertoldi, Brooke Cowan, Wade Shen, Christine Moran, Richard
  Zens, et~al. 2007.
\newblock Moses: Open source toolkit for statistical machine translation.
\newblock In \emph{Proc. ACL: Demos}, pages 177--180.

\bibitem[{Ladner and Fischer(1980)}]{ladner+fischer:jacm80}
Richard~E. Ladner and Michael~J. Fischer. 1980.
\newblock \href {https://doi.org/10.1145/322217.322232} {Parallel prefix
  computation}.
\newblock \emph{J. ACM}, 27(4):831--838.

\bibitem[{Li and Talwalkar(2019)}]{li2019random}
Liam Li and Ameet Talwalkar. 2019.
\newblock Random search and reproducibility for neural architecture search.

\bibitem[{Maclaurin et~al.(2015)Maclaurin, Duvenaud, and
  Adams}]{maclaurin2015gradient}
Dougal Maclaurin, David Duvenaud, and Ryan Adams. 2015.
\newblock Gradient-based hyperparameter optimization through reversible
  learning.
\newblock In \emph{Proc. ICML}, pages 2113--2122.

\bibitem[{Mauro et~al.(2012)Mauro, Christian, and Marcello}]{mauro2012wit3}
Cettolo Mauro, Girardi Christian, and Federico Marcello. 2012.
\newblock Wit3: Web inventory of transcribed and translated talks.
\newblock In \emph{Proc. EAMT}, pages 261--268.

\bibitem[{Murray and Chiang(2015)}]{murrayauto}
Kenton Murray and David Chiang. 2015.
\newblock Auto-sizing neural networks: With applications to $n$-gram language
  models.
\newblock In \emph{Proc. EMNLP}.

\bibitem[{Papineni et~al.(2002)Papineni, Roukos, Ward, and
  Zhu}]{papineni2002bleu}
Kishore Papineni, Salim Roukos, Todd Ward, and Wei-Jing Zhu. 2002.
\newblock {BLEU}: a method for automatic evaluation of machine translation.
\newblock In \emph{Proc. ACL}, pages 311--318.

\bibitem[{Parikh and Boyd(2014)}]{parikh+boyd:2014}
Neal Parikh and Stephen Boyd. 2014.
\newblock Proximal algorithms.
\newblock \emph{Foundations and Trends in Optimization}, 1(3):123--231.

\bibitem[{Popel and Bojar(2018)}]{popel2018training}
Martin Popel and Ond{\v{r}}ej Bojar. 2018.
\newblock Training tips for the {T}ransformer model.
\newblock \emph{The Prague Bulletin of Mathematical Linguistics},
  110(1):43--70.

\bibitem[{Press and Wolf(2017)}]{press-wolf:2017:EACLshort}
Ofir Press and Lior Wolf. 2017.
\newblock \href {http://www.aclweb.org/anthology/E17-2025} {Using the output
  embedding to improve language models}.
\newblock In \emph{Proc. EACL: Volume 2, Short Papers}, pages 157--163.

\bibitem[{Quattoni et~al.(2009)Quattoni, Carreras, Collins, and
  Darrell}]{quattoni2009efficient}
Ariadna Quattoni, Xavier Carreras, Michael Collins, and Trevor Darrell. 2009.
\newblock An efficient projection for $l_{1,\infty}$ regularization.
\newblock In \emph{Proc. ICML}, pages 857--864.

\bibitem[{Sennrich and Haddow(2016)}]{sennrich2016linguistic}
Rico Sennrich and Barry Haddow. 2016.
\newblock Linguistic input features improve neural machine translation.
\newblock In \emph{Proc. First Conference on Machine Translation: Volume 1,
  Research Papers}, volume~1, pages 83--91.

\bibitem[{Sennrich et~al.(2016)Sennrich, Haddow, and
  Birch}]{sennrich-haddow-birch:2016:P16-12}
Rico Sennrich, Barry Haddow, and Alexandra Birch. 2016.
\newblock \href {http://www.aclweb.org/anthology/P16-1162} {Neural machine
  translation of rare words with subword units}.
\newblock In \emph{Proc. ACL}, pages 1715--1725.

\bibitem[{Snoek et~al.(2012)Snoek, Larochelle, and Adams}]{snoek2012practical}
Jasper Snoek, Hugo Larochelle, and Ryan~P. Adams. 2012.
\newblock Practical {B}ayesian optimization of machine learning algorithms.
\newblock In \emph{Advances in Neural Information Processing Systems}, pages
  2951--2959.

\bibitem[{Strubell et~al.(2019)Strubell, Ganesh, and
  McCallum}]{strubell2019energy}
Emma Strubell, Ananya Ganesh, and Andrew McCallum. 2019.
\newblock Energy and policy considerations for deep learning in {NLP}.
\newblock In \emph{Proc. ACL}.

\bibitem[{Sutskever et~al.(2014)Sutskever, Vinyals, and
  Le}]{sutskever2014sequence}
Ilya Sutskever, Oriol Vinyals, and Quoc~V Le. 2014.
\newblock Sequence to sequence learning with neural networks.
\newblock In \emph{Advances in Neural Information Processing Systems}, pages
  3104--3112.

\bibitem[{Vaswani et~al.(2017)Vaswani, Shazeer, Parmar, Uszkoreit, Jones,
  Gomez, Kaiser, and Polosukhin}]{vaswani2017attention}
Ashish Vaswani, Noam Shazeer, Niki Parmar, Jakob Uszkoreit, Llion Jones,
  Aidan~N. Gomez, {\L}ukasz Kaiser, and Illia Polosukhin. 2017.
\newblock Attention is all you need.
\newblock In \emph{Advances in Neural Information Processing Systems}, pages
  5998--6008.

\bibitem[{Vose et~al.(2019)Vose, Balma, Heye, Rigazzi, Siegel, Moise, Robbins,
  and Sukumar}]{vose2019}
Aaron Vose, Jacob Balma, Alex Heye, Alessandro Rigazzi, Charles Siegel, Diana
  Moise, Benjamin Robbins, and Rangan Sukumar. 2019.
\newblock Recombination of artificial neural networks.
\newblock {arXiv}:1901.03900.

\end{thebibliography}
\bibliographystyle{acl_natbib}

\end{document}